\documentclass[10pt,twocolumn,letterpaper]{article}

\usepackage{cvpr}              % To produce the CAMERA-READY version
\usepackage{lipsum}
\usepackage{tikz}
\usepackage[latin1]{inputenc}
\usepackage{pgfplots}
\usetikzlibrary{backgrounds,scopes}   %<------- Load libraries
\pgfplotsset{compat=1.3}
\usetikzlibrary{shadings,spy,calc}
\usetikzlibrary {arrows.meta} 
\usepackage{bm}
\usepackage{colortbl}
\usepackage{booktabs} 
\usepackage{graphicx} % 
\usepackage{multirow} % add
\usepackage{bm}

\definecolor{cvprblue}{rgb}{0.21,0.49,0.74}
\usepackage[pagebackref,breaklinks,colorlinks,allcolors=cvprblue]{hyperref}

\def\paperID{12602} % *** Enter the Paper ID here
\def\confName{CVPR}
\def\confYear{2025}

\title{Learnable Infinite Taylor Gaussian for Dynamic View Rendering}

\author{Bingbing Hu$^{1}$\thanks{These authors contributed equally to this work}
\hspace{4pt}Yanyan Li$^{2}$\footnotemark[1] 
\hspace{4pt}Rui Xie$^{1}$ 
\hspace{4pt}Bo Xu$^{2}$ 
\hspace{4pt}Haoye Dong$^{2}$ 
\hspace{4pt}Junfeng Yao $^{1,3}$\thanks{Corresponding author}
\hspace{4pt}Gim Hee Lee$^{2}$ 
\and
$^{1}$ School of Film, School of Informatics, Xiamen University \\
$^{2}$ National University of Singapore \\
$^{3}$ Key Laboratory of Digital Protection and Intelligent Processing of \\ Intangible Cultural Heritage of Fujian and Taiwan Ministry of Culture and Tourism \\
}

\begin{document}

\maketitle

\begin{abstract}
Capturing the temporal evolution of Gaussian properties such as position, rotation, and scale is a challenging task due to the vast number of time-varying parameters and the limited photometric data available, which generally results in convergence issues, making it difficult to find an optimal solution. While feeding all inputs into an end-to-end neural network can effectively model complex temporal dynamics, this approach lacks explicit supervision and struggles to generate high-quality transformation fields. On the other hand, using time-conditioned polynomial functions to model Gaussian trajectories and orientations provides a more explicit and interpretable solution, but requires significant handcrafted effort and lacks generalizability across diverse scenes. 
To overcome these limitations, this paper introduces a novel approach based on a learnable infinite Taylor Formula to model the temporal evolution of Gaussians. This method offers both the flexibility of an implicit network-based approach and the interpretability of explicit polynomial functions, allowing for more robust and generalizable modeling of Gaussian dynamics across various dynamic scenes.
Extensive experiments on dynamic novel view rendering tasks are conducted on public datasets, demonstrating that the proposed method achieves state-of-the-art performance in this domain. More information is available on our
project page (\url{https://ellisonking.github.io/TaylorGaussian}).
% The goal of capturing how the properties (such as position, rotation, and scale) of each Gaussian evolve over time is challenging, as the vast number of time-varying Gaussian parameters is constrained by limited photometric data. This often leads to convergence in different directions, making it difficult to guarantee finding an optimal minimum. 
% % 
% Feeding all inputs into an end-to-end network is an acceptable choice, as it enables the model to learn the complex temporal dynamics and interdependencies between Gaussian properties directly from the data. However, the drawback is that the process cannot be explicitly supervised, and the network is struggle to produce high-quality transformation fields. 
% %
% Compared to the implicit representation of Gaussians, time-conditioned polynomial functions for modeling trajectories and orientations offer a more explicit approach. However, the disadvantage is that they require more handcrafted effort to define the complexity of the approximating function, making it difficult to develop a generalizable model that works across different types of scenes.
% To address the critical challenges in this domain, this paper proposes a novel approach by establishing a \textbf{learnable infinite Taylor series} to model this process. 
\end{abstract}
\section{Introduction}
\label{sec:intro}

% \begin{figure}
%   \centering
%   \includegraphics[width=\linewidth]{method_comparison_grid.png}%{reconstruction-fig.jpg} 
%   \caption{Comparison of Methods based on the mean values of SSIM and PSNR on the Neural 3D Video Dataset~\cite{an15}.}
%   \label{fig:method_comparison}
% \end{figure}

% \begin{figure*}[htbp]
%   % \centering
%   \includegraphics[width=\textwidth]{method_comparison2.png} % 替换为你的图像文
%   \caption{Comparison of Methods Based on the Mean Values of SSIM and PSNR.}
%   \label{fig:fig1}
%   \Description{}
% \end{figure*}

% \begin{figure*}[htbp]
%   % \centering
%   \includegraphics[width=\textwidth]{reconstruction-fig.jpg} % 替换为你的图像文
%   \caption{The quality of our method is demonstrated on the Neural 3D Video (N3DV) dataset.}
%   \label{fig:fig1}
%   \Description{}
% \end{figure*}
% 插入跨栏图形
% \begin{figure}[]
%   \centering
%   \includegraphics[width=\linewidth]{meetroom-recon.jpg}  %{reconstruction-fig.jpg} % 替换为你的图像文件名
%   \caption{The quality of our method is demonstrated on the MeetRoom dataset.}
%   \label{fig:reconstruction-fig}
% \end{figure}

Recently, 3D Gaussian Splatting (3DGS) has achieved groundbreaking progress in dynamic scene reconstruction~\cite{zhang2024street,a2,a11}, especially with the adoption of tile-based rasterization techniques as a replacement for traditional volumetric rendering methods~\cite{a1,an7,an8,an9,lin2024unsupervised,hou2025mvgsr}. This innovation has garnered substantial attention within the academic community. Many researchers have started to leverage 3DGS for 4D scene reconstruction~\cite{a2,a3,a4,a5}, aiming to accurately capture and model the evolving 3D structure and appearance of scenes over time, enabling novel view synthesis at arbitrary time points. Although modeling static scenes has seen significant advances~\cite{a1,a6,a7,zhu2024robust,fan2024instantsplat}, dynamic scene reconstruction remains challenging due to factors such as the complexity of object motion, topological changes, and spatial or temporal sparsity in observations~\cite{a5,a9,a10,an8,an7}. These issues make accurate reconstruction of dynamic scenes a technical challenge, requiring ongoing research and innovation to overcome.

\begin{figure*}[]
  \centering
  \includegraphics[width=1.0\textwidth]{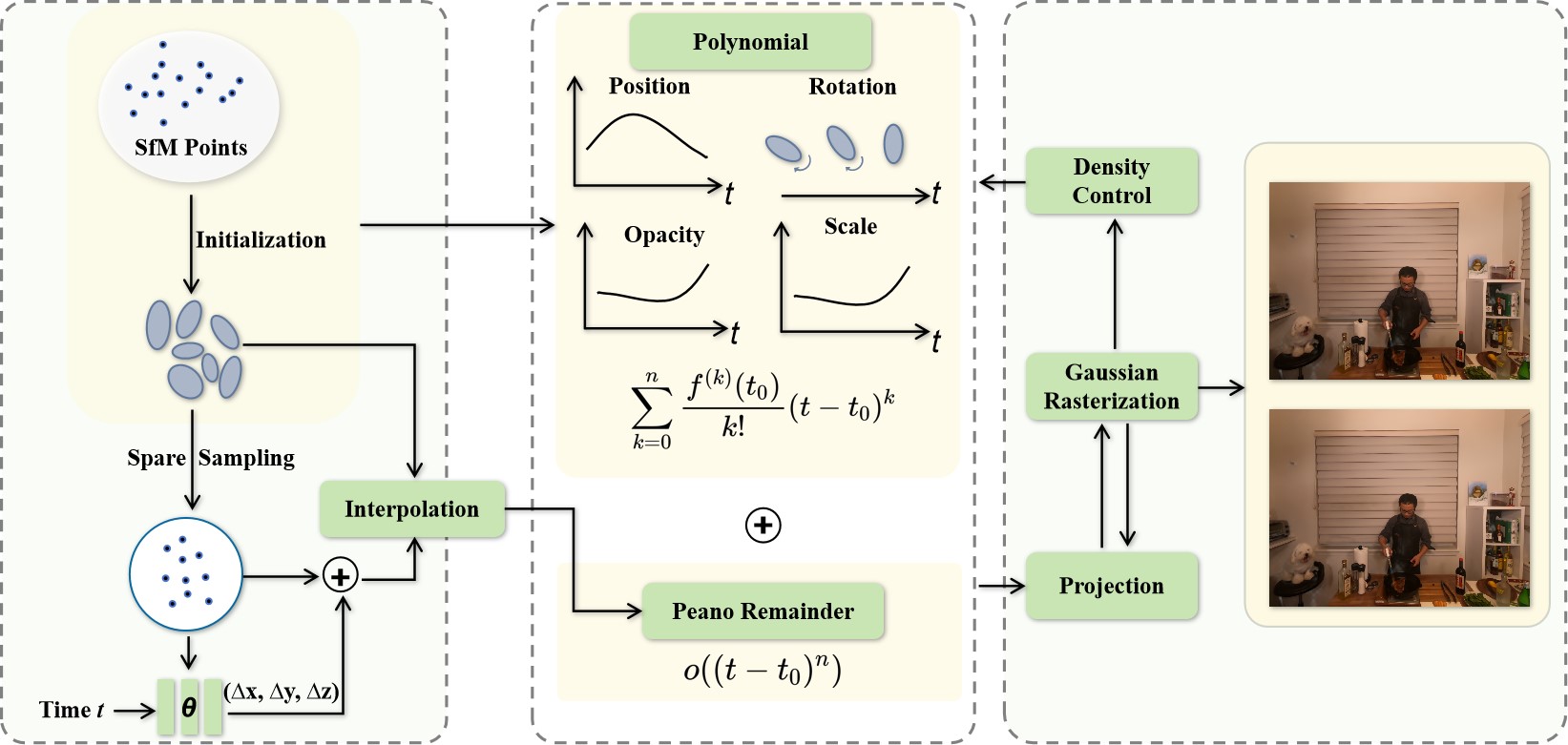} % 替换为你的图像文件名
  \caption{The detailed architecture of the proposed method. The framework includes Gaussian Initialization, Sparse Point Sampling, Gaussian Point Interpolation, and Gaussian Transformation Fields Modeling.}
  \label{fig:Dynamic Gaussian Splatting}
  %\Description{}
\end{figure*}

Extending static 3DGS techniques to continuous representations of dynamic scenes is a challenging task. Some researchers have explored various approaches~\cite{a8,a11,a12} to address this challenge. In the representation and rendering of dynamic scenes,  deformable 3DGS (D3DGS)~\cite{a3} introduces deformation fields to simulate dynamic changes, yet issues with continuity and frame-to-frame correlation affect reconstruction quality. The  Streaming Radiance Fields (StreamRF)~\cite{an7} propose an efficient dynamic scene reconstruction method using an explicit grid-based approach, synthesizing 3D video through an incremental learning paradigm and a narrow-band optimization strategy. 
Representing and rendering dynamic scenes has always been an important and challenging task. In dynamic scenes, many parameters of the Gaussian functions change over time. However, due to the limitations of available photometric data, it is difficult for models to accurately learn the complex temporal dynamics and interdependencies between Gaussian function attributes. This challenge becomes even more pronounced when simulating complex motion.
To address the challenge of capturing the temporal evolution of Gaussian motion attributes, we propose an innovative 3DGS framework. As shown in Figure ~\ref{fig:Dynamic Gaussian Splatting}, this framework deeply explores the mathematical principles behind Gaussian point motion and analyzes their trajectories to accurately simulate complex motion dynamics. Specifically, we introduce a learnable infinite Taylor series to model the motion trajectories of Gaussian points in dynamic scenes. By tracking the evolution of Gaussian points over time, we can precisely capture key attributes (such as position, opacity, and scale) at each time step. This approach not only provides a solid mathematical foundation for 3D reconstruction and view synthesis but also offers a novel perspective for dynamic scene modeling.
% To tackle this problem, we propose an innovative 3DGS framework. This framework initiates the 3D reconstruction process by performing precise sampling within Gaussian points and training an initial motion field specifically for dynamic scenes, establishing refined spatiotemporal proximity among Gaussian points. We delve into the mathematical principles underlying Gaussian point motion, analyzing their trajectories to accurately characterize the complex evolution of motion, thereby significantly enhancing the fidelity and quality of 3D reconstructions. 
% Additionally, we designed a time-varying segmented learning rate strategy to markedly improve model robustness throughout the training process. This strategy finely adjusts the learning rate to balance the stability required in early training with detailed optimization in later stages, ensuring that the model maintains robustness while achieving efficient convergence across the entire training cycle.

Our main contributions are as follows:
\begin{itemize}
    \item  A novel perspective, learnable infinite Taylor Formula, is proposed to model the transformation fields of dynamic Gaussian primitives over time.
    \item The dominant component of our transformation field is modeled using a third-order Taylor expansion to achieve large motion estimation.
    \item  The Peano remainder is constructed via the deformation field, forming a complete Taylor series to estimate the motion model without approximation.
 \item Extensive experiments show that our method outperforms the baseline in both qualitative and quantitative multi-view evaluations, enabling more accurate and faithful modeling of dynamic content.
\end{itemize}

\section{Related Work}
\label{sec:related}

\paragraph{Dynamic Novel View Synthesis.}
In the field of dynamic free-viewpoint rendering, multi-view video inputs are commonly used. Before the advent of more advanced techniques such as  Neural Radiance Field (NeRF)~\cite{an1} and Gaussian Splatting~\cite{a1}, very few works tackled this problem but rather the static version of the issue since the cost of utilizing traditional volumetric rendering techniques is too computationally expensive. Most approaches use traditional volumetric rendering techniques without much space optimization. Neural Volumes~\cite{an12} is a pioneering work that employs an encoder-decoder network to convert images into 3D volumes. The volumes are rendered with intricate details using volumetric techniques. However, it does not achieve resolutions similar to traditional textured mesh surfaces.

\paragraph{Dynamic NeRF.}
% To tackle the problem of low fidelity of previous novel view synthesis methods, NeRF was introduced~\cite{an1}. NeRF~\cite{an1} leverages a multilayer perceptron (MLP) to implicitly model static scenes, inspiring subsequent works to explore its extension to model dynamic scenes. Some approaches employ deformation fields to estimate scene motion, warping the 3D radiance field of each frame to one or more canonical frames. Innovative methods like NeRF-W~\cite{an13} and Block-NeRF~\cite{an14} capture diverse scene appearance details through image embeddings and accurately reconstruct scenes from in-the-wild image collections. Additionally, significant progress has been made in extending static neural rendering techniques to dynamic scenes.

Dynamic NeRF (DyNeRF)~\cite{an15}, for instance, trains a NeRF for dynamic scenes using a straightforward neural network structure. It takes 3D positions and time as inputs and employs a series of fully connected neural networks to predict properties such as color and density. By performing temporal interpolation on intermediate features, DyNeRF~\cite{an15} enhances its capacity to represent dynamic features while maintaining structural simplicity. 
 {Mixed Neural Voxels (MixVoxels)~\cite{an16} accelerates the rendering process by blending static and dynamic voxels. NeRFPlayer~\cite{an20} intricately decomposes the scene into static, newly added, and deformed fields, introducing an innovative feature flow channel concept. Techniques such as K-Planes~\cite{an19}, HexPlane~\cite{an4}, and Tensor4D~\cite{an21} decompose the 4D spatiotemporal domain into 2D feature planes which optimizes model size. HyperNeRF~\cite{an22} combines per-frame appearance and deformation embeddings, further enhancing expressiveness. 
% Sync-NeRF~\cite{an17} introduces time offsets to correct temporal embedding misalignments in dynamic scenes from unsynchronized video. It further encodes the evolution of each Gaussian value over time with Gaussian latent embeddings to capture the unique state of each frame through time embeddings.
Additionally, several methods~\cite{a9, an24, an25} model dynamic scenes as 4D radiance fields; however, these approaches face high computational costs due to the complexity of ray-point sampling and volumetric rendering.

% [7] Chen Gao, Ayush Saraf, Johannes Kopf, and Jia-Bin Huang.Dynamic view synthesis from dynamic monocular video. In ICCV, 2021. 2
% [8] Lin Gao, Yu-Kun Lai, Jie Yang, Ling-Xiao Zhang, Shihong Xia, and LeifKobbelt. Sparse data driven mesh deformation.IEEE TVCG, 27(3):2085–2100, 2019. 2
% [35] Sungheon Park, Minjung Son, Seokhwan Jang, Young Chun Ahn, Ji-Yeon Kim, and Nahyup Kang. Temporal interpolation is all you need for dynamic neural radiance fields. In
% CVPR, 2023. 2

\paragraph{Dynamic Gaussian Splatting.}
Inspired by 3DGS~\cite{a1}, dynamic 3D Gaussian technology extends the fast rendering capabilities of 3DGS to dynamic scene reconstruction. 4D Gaussian splatting (4DGS)~\cite{a2} introduces a novel explicit representation that combines 3D Gaussians with 4D neural voxels, proposing a decomposition neural voxel encoding algorithm inspired by HexPlane~\cite{an4} to efficiently construct Gaussian features from 4D neural voxels. A lightweight MLP is then applied to predict Gaussian deformations at new timestamps. D3DGS~\cite{a3} presents a deformable 3DGS framework for dynamic scene modeling, where time is conditioned on the 3DGS. The learning process is transformed into a canonical space, where a purely implicit deformable field is jointly trained with the learnable 3DGS, resulting in a time-independent 3DGS, decoupling motion from geometric structure. 3D Gaussians for Efficient Streaming (3DGStream)~\cite{a13} enables efficient streaming of photo-realistic Free-Viewpoint Videos (FVVs) for dynamic scenes, leveraging a compact Neural Transformation Cache (NTC) to simulate the translation and rotation of  3D Gaussians. This significantly reduces the training time and storage space required for each frame of FVVs, while introducing an adaptive  3D Gaussians addition strategy to handle new objects in dynamic scenes. 

% The dynamic 3D Gaussian technique extends the rapid rendering capabilities of 3D Gaussian splatting to dynamic scene reconstruction. In this field, D3DGS~\cite{a3} employs an implicit function to simultaneously handle the temporal and spatial dimensions of Gaussian distributions. 4DGS~\cite{a11} further refines this approach by decomposing four-dimensional Gaussians into time-conditioned 3D Gaussians and a marginal 1D Gaussian, capturing dynamic changes with greater precision. 4DGaussians~\cite{a2} decodes features from a multi-resolution HexPlane~\cite{an4}, enabling temporal deformation of 3D Gaussian distributions. STG~\cite{a12} represents the evolution of 3D Gaussian values over time by incorporating temporal opacity and polynomial functions for each Gaussian value.
    
\section{Preliminaries}
\label{subsec:prelim}

\subsection{3D Gaussian Splatting}\label{subsec:3DGS}
Given a complete 3D covariance matrix \(\bm{\Sigma}\) and a mean vector \(\bm{\mu}\) in the world coordinate frame, the 3D Gaussian distribution can be defined as:

\begin{equation}
    \mathcal{G}(\mathbf{x} | \bm{\mu}, \bm{\Sigma}) = e^{-\frac{1}{2} (\mathbf{x}-\bm{\mu})^{T} \bm{\Sigma}^{-1} (\mathbf{x}-\bm{\mu})} 
\end{equation}
where $\bm{\mu}\in \mathbb{R}^3$, $\bm{\Sigma} \in \mathbb{R}^{3\times 3}$. To ensure that the covariance matrix is semi-positive definite, it is represented using a diagonal scaling matrix \(\mathbf{S}_i = \text{Diag}[s_1 \; s_2 \; s_3] \in \mathbb{R}^{3 \times 3}\) and a rotation matrix \(\mathbf{R} \in SO(3)\). This can be expressed as:

\begin{equation}
    \bm{\Sigma} = \mathbf{R} \mathbf{S} (\mathbf{S})^{\intercal} (\mathbf{R})^{\intercal} 
\end{equation}
where \(SO(3)\) denotes the special orthogonal group. In addition to the position and shape parameters, spherical harmonics coefficients \(\mathbf{C} \in \mathbb{R}^{(m+1)2 \times 3}\) (where \(m\) is the degree of freedom) and opacity \(\alpha \in \mathbb{R}\) also play important roles in rendering the colored image.

The color of a target pixel can be synthesized by splatting and blending these \(N\) organized Gaussian points that overlap with the pixel. First, the splatting operation forms 2D Gaussians \(\mathcal{N}(\bm{\mu}_I, \bm{\Sigma}_I)\) on the image plane from the 3D Gaussians \(\mathcal{N}(\bm{\mu}_w, \bm{\Sigma}_w)\) in the world coordinates based on the camera poses. Specifically:
\begin{equation}
    \bm{\mu}_I = \Pi (\mathbf{T}_{cw} \bm{\mu}_w), \; \bm{\Sigma}_I = \mathbf{J} \mathbf{W}_{cw} \bm{\Sigma}_w \mathbf{W}_{cw}^{\intercal} \mathbf{J}^{\intercal}
\end{equation}

where \(\mathbf{T}_{cw} \in SE(3)\) is the camera pose, representing the transformation from the world coordinate to the camera coordinate in the special Euclidean group. The components \(\mathbf{W}_{cw}\) and \(\mathbf{T}_{cw}\) represent the rotation and translation, respectively. \(\mathbf{J}\) is the Jacobian matrix of the affine approximation of the projective transformation~\cite{li2024smilesplat,zwicker2001ewa}. Therefore, the blending operation is then given by:

\begin{equation}
    \mathcal{C}_{\mathbf{p}} = \sum_{i \in N} c_i \alpha_i \prod_{j=1}^{i-1} (1 - \alpha_j)
\end{equation}
where \(c_i\) and \(\alpha_i\) represent the color and opacity of the \(i\)-th point, respectively.

\subsection{Representation of Dynamic Gaussian}

In contrast to the 3D Gaussian representation introduced in Section~\ref{subsec:3DGS}, we incorporate a timestamp \( t \) into each Gaussian, resulting in a 4D Gaussian representation: 
\begin{equation}
    \mathcal{G}^{4D} = [\bm{\mu} \; \bm{\Sigma} \; c \; o \; t].    
\end{equation}

Inspired by the work of~\cite{a12,a1,chen2023neurbf}, we model the opacity of the \(i^{th}\) 4D Gaussian \(\mathcal{G}^{4D}_i\) as a time-dependent function, defined as follows:

\begin{eqnarray}
\alpha_{i}(t) = \bm{\sigma}_{i}(t) e^{\left(-\frac{1}{2} \left( \mathbf{x} - \bm{\mu}_{i}(t) \right)^{T} \Sigma_{i}(t)^{-1} \left( \mathbf{x} - \bm{\mu}_{i}(t) \right) \right)}
\end{eqnarray}
where \(\sigma_{i}(t)\) represents the time-dependent opacity of the Gaussian, and \(\bm{\mu}_{i}(t)\) and \(\Sigma_{i}(t)\) are the position and covariance parameters, respectively, which evolve over time. Similar to how the mean and covariance are computed in the 1D Gaussian model, we estimate the temporal opacity \(\bm{\sigma}_{i}^{s}\) based on the following formulation:
\begin{equation}
\sigma_{i}(t) = \sigma_{i}^{s} e^{ -s_{i}^{\tau} \left| t - \bm{\mu}_{i}^{\tau} \right|^{2}}
\end{equation}
where \(\sigma_{i}^{s}\) is the stationary opacity (time-independent), \(s_{i}^{\tau}\) is a temporal scaling factor, and \(\exp(\cdot)\) represents the radial basis function (RBF)~\cite{chen2023neurbf}. Here, \(\bm{\mu}_{i}^{\tau}\) is the temporal center, and the expression models the decay of opacity over time.

\section{Theory}

\subsection{Fundamental Theory of Taylor Formula}
The geometric significance of the Taylor formula is that it uses polynomial functions to approximate the original function. Since polynomial functions can be differentiated to any order, they are easy to compute and convenient for finding extrema or analyzing the properties of the function. Therefore, the Taylor formula provides valuable information about the a function model, which can be written as 
\begin{equation}
\begin{split}
    f(x) = &f(x_0) + f^{'}(x_0)(x-x_0) + \frac{1}{2!} f^{''}(x_0)(x-x_0)^2  \\
     &+ \frac{1}{3!} f^{(3)}(x_0)(x-x_0)^3 + \dots  \\
    & +\frac{1}{n!}f^{(n)}(x_0) (x-x_0)^n + R_n(x)
\end{split}
\end{equation}
here the notation \( n! \) refers to the factorial of \( n \). The function \( f^{(n)}(\cdot) \) represents the \( n^{th} \) derivative of \( f \) evaluated at the point \( x_0 \). The derivative of order zero of \( f \) is simply \( f \) itself, and both \( (x - x_0)^0 \) and \( 0! \) are defined as $1$. And the remainder term $R_n(x)$ can be defined in Peano's form, which can be described as 
\begin{equation}
    R_n(x) = o_n(x)(x-x_0)^n
\end{equation}
where $\lim_{x\to x_0}o_n(x) = 0$.

Based on the Taylor formula, Taylor series is a method of expanding a function $f(x)$ into a sum of powers, with the aim of approximating a complex function using relatively simple functions, which can be expressed as 
\begin{equation}
    f(x) \approx \sum_{k=0}^{n} c_k (x-x_0)^k
\end{equation}
where it means the function can be established an approximation
via several simpler polynomial functions. We have to note that error analysis must be provided to assess the reliability of the approximation during the process.

\subsection{How to Approximate the Moving Functions of 4D Gaussians?}

The goal of capturing how the properties (such as position, rotation, and scale) of each Gaussian evolve over time is challenging, as the vast number of time-varying Gaussian parameters is constrained by limited photometric data. This often leads to convergence in different directions, making it difficult to guarantee finding an optimal minimum. 

Feeding all inputs into an end-to-end network is a acceptable choice, as it enables the model to learn the complex temporal dynamics and interdependencies between Gaussian properties directly from the data. However, the drawback is that the process cannot be explicitly supervised, and the network is struggle to produce high-quality transformation fields. 
Compared to the implicit representation of Gaussians, time-conditioned polynomial functions for modeling trajectories and orientations offer a more explicit approach. However, the disadvantage is that they require more handcrafted effort to define the complexity of the approximating function, making it difficult to develop a generalizable model that works across different types of scenes.

To address the critical challenges in this domain, this paper proposes a novel approach by establishing a \textbf{learnable infinite Taylor series} to model this process. 
To be specific, we track the movement of Gaussian points over time and use Taylor Formula to capture key attributes such as position, rotation, and scale at different timestamps, where the formula is decomposed into two components. The first component applies the Taylor expansion to construct polynomials $f_k(t)$ that approximate large-scale transformations, while the second uses an end-to-end neural network to learn the Peano remainder $\mathcal{H}_k(t)$ term. With this carefully designed approach, the proposed method constructs a complete Taylor series that estimates the motion model without relying on approximations. Therefore, the theory of the proposed method can be expressed as:
\begin{equation}
    \mathcal{T}_i (t) = f_k(t) + \mathcal{H}_k(t)
\end{equation}
where $\mathcal{T}$ denotes the spatial transformation of the Taylor Gaussian at timestamp $t$. In following section~\ref{subsec:Taylor_1} and~\ref{subsec:Taylor_2}, we will introduce the details of strategy to estimate Taylor Gaussian.  

\subsection{Taylor Expansion of Transformation Field Modeling}
\label{subsec:Taylor_1}

% In this section, the proposed method for enriching local Gaussian primitives is introduced in this section. Inspired by~\cite{SCGS,4dgs},  we first estimate the transformation of each local Gaussian at timestamp $t$ based on the linear blend skinning (LBS)~\cite{} and time-dependent deep motion embedding trained in Section~\ref{sec:gp}. Specifically, the k-neighboring GPs of each Gaussian is detected and then they are jointly considered to compute via the RBF algorithm~\cite{}:
% \begin{eqnarray}
%    scgs: EQ ---> 5
% \end{eqnarray}
% \paragraph{9-DoF Gaussian Pose Modeling.}

% In this section, the dominate component of our transformation field is modeled based on the third-order Taylor expansion. And the goal of this transformation is estimate the time-dependent position, scaling, and orientation.

% In this section, the main components of the Taylor series are modeled using a third-order Taylor expansion, with the aim of estimating the time-dependent position, scale, and orientation.

In this section, the dominant component of our transformation field is modeled using a third-order Taylor expansion, and the goal of this transformation field is to estimate the time-dependent position, scaling, and orientation.

\textit{Position Motion.} To model the position of a Gaussian at different timestamps, we use a time-dependent polynomial function to describe its smooth trajectory:
\begin{equation}
    % \bm{\mu}_i(t) = \sum_{k=0}^{n_p} b_{i,k} (t - \mu_i^\tau)^k,
    \bm{p}_i(t) = \sum_{k=0}^{n} \frac{1}{k!} f_{p}^{(k)}\left(t_{\tau} \right) (t -t_{\tau})^k
\end{equation}
where \(\bm{p}_i(t)\) represents the position of \(\mathcal{G}_i\) at time \(t\), \(k\) denotes the order, and \(f_{p}^{(k)}\) represents the \(k\)-th derivative of \(f_{p}\). For \(\frac{1}{k!} f_{p}^{(k)}(t_{\tau})\), where \(\frac{1}{k!} f_{p}^{(k)}(t_{\tau}) \in \mathbb{R}\), \(f_{p}^{(k)}(t_{\tau})\) represents the \(k\)-th derivative of the Taylor series of \(f_{p}\) evaluated at \(t_{\tau}\), with \(t_{\tau}\) being the time center.
% where \(\bm{p}_i(t)\) represents the position of \(\mathcal{G}_i\) at time \(t\), \(k\) denotes the order, and \( f_{p}^{(k)} \) represents the \(k\)-th derivative of \(f\). For \(\frac{1}{k!} f_{p}^{(k)}(t_{\tau})\), where \(t_{\tau}\) is the time center, \(\frac{1}{k!} f_{p}^{(k)}(t_{\tau}) \in \mathbb{R}\) is the \(k\)-th derivative of the Taylor series $f_{p}$ evaluated at \(t_{\tau}\).

% where \( \mu_i(t) \) represents the position of \( \mathcal{G}_i \) at time \( t \), and \( \{b_{i,k}\}_{k=0}^{n_p} \), with \( b_{i,k} \in \mathbb{R} \), are the coefficients of the polynomial. \( \mu_i^\tau \) is the temporal center, and \( k \) denotes the order of the polynomial.

\textit{Scaling Consistency.}
During the motion, the scale vector of each Gaussian is assumed to change smoothly. Therefore, we model this scaling behavior as follows:
\begin{equation}
    % \mathbf{s}_i(t) = \sum_{k=0}^{n_s} s_{i,k} (t - \mu_i^\tau)^k,    
    \bm{s}_i(t) = \sum_{k=0}^{m} \frac{1}{k!} f_{s}^{(k)}\left(t_{\tau} \right) (t - t_{\tau})^k  
\end{equation}
where \(\bm{s}_i(t)\) represents the scale of \(\mathcal{G}_i\) at time \(t\). For \(\frac{1}{k!} f_{s}^{(k)}(t_{\tau})\), where \(\frac{1}{k!} f_{s}^{(k)}(t_{\tau}) \in \mathbb{R}\), \(f_{s}^{(k)}(t_{\tau})\) represents the \(k\)-th derivative of \(f_{s}\) at the time center \(t_{\tau}\), and \(k!\) denotes the factorial of \(k\).
% where \( \bm{s}_i(t) \) represents the scale at time \( t \), and \( \{s_{i,k}\}_{k=0}^{n_s} \), with \( s_{i,k} \in \mathbb{R} \), are the polynomial coefficients.

\textit{Orientation Motion.}
% For modeling the orientation motion, we use quaternion representation and apply a polynomial function \( \mathbf{q}_i(t) \) to accurately capture the continuous rotational motion of the object. 
For modeling the orientation motion, we use quaternion representation and apply a Taylor expansion to \( \mathbf{q}_i(t) \) to accurately capture the continuous rotational motion of the object. 
This approach enables us to effectively model the rotational dynamics over time, leading to more precise orientation control in dynamic scene reconstruction. It not only improves the accuracy of rotational motion description but also enhances the flexibility and adaptability of the model in handling complex dynamic changes:

\begin{equation}
    % \mathbf{q}_i(t) = \sum_{k=0}^{n_q} c_{i,k} (t - \mu_i^\tau)^k,
    \bm{q}_i(t) = \sum_{k=0}^{l} \frac{1}{k!} f_{q}^{(k)}\left(t_{\tau} \right) (t - t_{\tau})^k
\end{equation}
where \(\mathbf{q}_i(t)\) represents the Taylor expansion of the rotation at time \(t\). In this expansion, \(\frac{1}{k!} f_{q}^{(k)}(t_{\tau}) \in \mathbb{R}\), where \(f_{q}^{(k)}(t_{\tau})\) is the \(k\)-th derivative of \(f_q\) at the time center \(t_{\tau}\), and \(k!\) is the factorial of \(k\). This expression captures the local variation of \(f_q\) around \(t_{\tau}\) and is essential for constructing the Taylor series to approximate \(f_q\) near \(t_{\tau}\).
\subsection{Peano Remainder of Transformation Fields Modeling}
\label{subsec:Taylor_2}

% Given Gaussians at timestamp $t_{j-1}, j\geq{2}$, we make use of a MLP $\phi$ to estimate the offset of each Gaussian at $t_j$ via 
% \begin{equation}
%     \left\{\begin{split}
%         \Delta \bm{\mu} &= \phi_{\mu} (f_d) \\
%         \Delta \mathbf{R} &= \phi_{\mathbf{R}} (f_d) \\
%         \Delta \mathbf{s} &= \phi_{s} (f_d) \\
%         \Delta o &= \phi_{o} (f_d) \\
%     \end{split} \right.
% \end{equation}
% Then the deformed Gaussian can be updated by $\mathcal{G}=[\mu+\Delta \bm{\mu}, \; \mathbf{R}\Delta \mathbf{R}, \; \mathbf{s}+\Delta\mathbf{s}]$.
In this section, the strategy of Peano Remainder estimation of the transformation fields is introduced in this section. First, we classify the Gaussians into two sets: Global Gaussian Primitives (GPs) and Local Gaussian Primitives (LPs). The GPs, which have global representative features, serve as the skeletons of objects, while the LPs play a critical role in achieving high-quality rendering. Specifically, $N$ GPs are initially selected from the Gaussian map using the farthest point sampling approach. Compared to the number of LPs, the number is much smaller, making the GPs sparse. Since GPs are assumed to remain stable across different views and time instances, we establish a time-dependent transformation prediction network to predict the translation and orientation of each GP in canonical coordinates.

In contrast to methods that estimate the temporal shifts of all Gaussian points through an MLP network~\cite{a2}, predicting shifts for all Gaussian points simultaneously often leads to weaker geometric and temporal consistency. To overcome this issue, we optimize the offsets of the GP points by using an MLP decoder to encode the features of the GPs. At time t, when querying each GP, the MLP provides the offset for that GP only via the following function:

\begin{equation}
     \Delta_{GP} = MLP(GP)
     % (\mathbf{R}_i^t, \mathbf{t}^t_i) = \mathcal{M}(\mathcal{F}, \mathcal{G}_g, t) 
    \label{eq:gp_eq}
\end{equation}

We then derive the Peano Remainder terms of the motion equation for LP points based on the GP deformation field at different time steps. The Peano Remainder for the LP points is interpolated using Linear Blend Skinning (LBS)~\cite{a14}. In many scenarios, the offset of a GP point influences the position of nearby LP points, meaning the offset of an LP point is constrained by the corresponding GP point's offset. As a result, the offsets of the LP points inferred from the GP points ensure spatial consistency (i.e., the positions between LP and GP points remain invariant, with nearby GP points unchanged) and temporal consistency (i.e., at the same time, LP points and their adjacent GP points exhibit consistent motion, maintaining rigidity between neighboring points) ~\cite{li2024st,huang2024sc}. We define a distance function $d_{ij}$ to represent the distance between a GP point  $G_i$ and an LP point $C_j$. The weight of neighboring GP points relative to the LP point is computed using the Gaussian-kernel RBF method~\cite{gao2019surfelwarp,dou2016fusion4d,newcombe2015dynamicfusion,huang2024sc}:
\begin{equation}
w_{i j}=\frac{\hat{w}_{i j}}{\sum_{j \in \mathcal{N}} \hat{w}_{i j}} \text {, where } \hat{w}_{i j}=\exp \left(-\frac{d_{ij}^{2}}{2 r_{j}^{2}}\right)
\end{equation}
here \( r_j \) is the learnable radius parameter for the GP point. The gradient descent method with backpropagation can learn the radius parameter.  $w_{ij}$ represents the weight of GP point $j$ to LP point $i$. The Peano remainder terms of the motion equation for LP points can be accurately estimated using LBS via the following function:

\begin{equation}
\Delta {\mu_{i}^{t}} =\sum_{j \in \mathcal{N}} w_{i j}\left(R_{j}^{t}\left(\mu_{i}-p_{j}\right)+p_{j}+\Delta d_{j}^{t}\right)
\end{equation}

\begin{equation}
\Delta q_{i}^{t} =\left(\sum_{j \in \mathcal{N}} w_{ij} r_{j}^{t}\right) \otimes q_{i}
\end{equation}
where \( R^t_j(t,j) \in \mathbb{R}^{3 \times 3} \) and \( r^t_j(t,j) \in \mathbb{R}^4 \) represent the predicted rotation matrix and quaternion representation at GP point \( j \), respectively, at time step \( t \). \( \Delta d^t_j \) denotes the offset by which \( p_j \) moves at time step \( t \). \( u_i \) represents the position of LP point \( i \), \( p_j \) represents the position of GP point \( j \), \( \Delta q_i^t \) represents the quaternion of LP point \( i \) at time step \( t \), and \( \Delta u_i^t \) represents the offset position of LP point \( i \) at time step \( t \).

% \begin{figure*}[htbp]
%   \centering
%   \includegraphics[width=\textwidth]{fig2-1.png} % 替换为你的图像文件名
%   \includegraphics[width=\textwidth]{fig2-2.png} % 替换为你的图像文
%   \begin{tikzpicture}
%     \node[font=\bf] at (11,-1.5) {};
%     \node[font=\bf] at (10,-1.5) {GT};
%     \node[font=\bf] at (6,-1.5) {Ours};
%     \node[font=\bf] at (1.5, -1.5) {4DGS~\cite{a2}};
%     \node[font=\bf] at (-3,-1.5) {D3DGS~\cite{a3}};
%   \end{tikzpicture}
%   \caption{Comparison of novel view rendering on the Neural 3D Video dataset, with problem regions highlighted in red boxes. More results are provided in the supplementary. }
%   \label{fig:fig2-1}
% \end{figure*}
\definecolor{lightred}{rgb}{1, 0.4, 0.4}

\begin{figure*}[htbp]
  \centering
  \includegraphics[width=\textwidth]{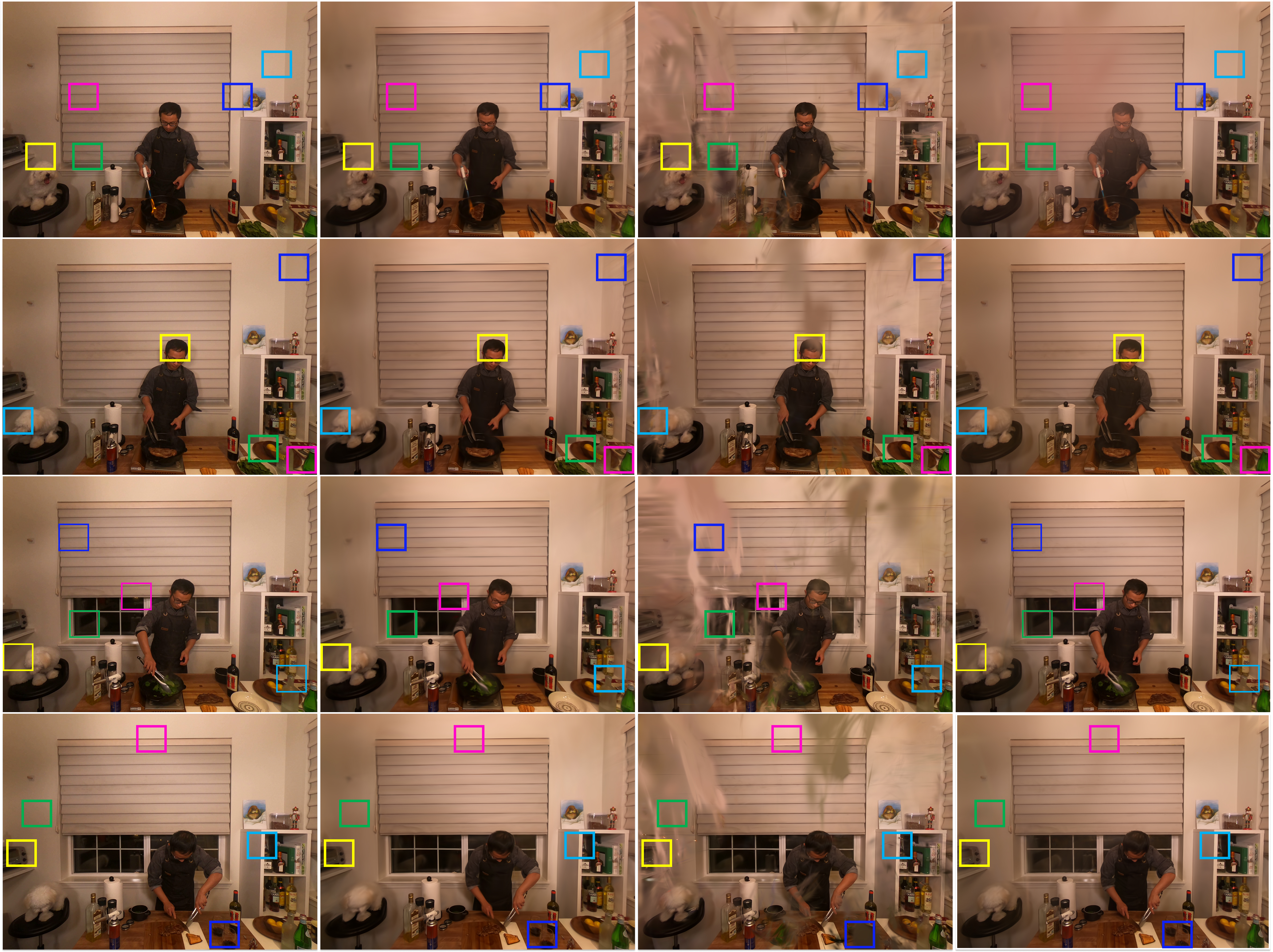} % 替换为你的图像文
  \begin{tikzpicture}
    \node[font=\bf] at (11,-1.5) {};
    \node[font=\bf] at (10,-1.5) {4DGS~\cite{a2}};
    \node[font=\bf] at (6,-1.5) {D3DGS~\cite{a3}};
    \node[font=\bf] at (1.5, -1.5) {Ours};
    \node[font=\bf] at (-3,-1.5) {GT};

  \end{tikzpicture}
  \caption{Comparison of novel view rendering on the N3DV dataset, with problem regions highlighted in boxes. More results can be found in the supplementary material and on our project website. }
  \label{fig:fig2-1}
\end{figure*}

% \begin{figure*}[htbp]
%   \centering
%   \includegraphics[width=\textwidth]{fig2-2.png} % 替换为你的图像文
%   \begin{tikzpicture}
%     \node[font=\bf] at (11,-1.5) {};
%     \node[font=\bf] at (10,-1.5) {GT};
%     \node[font=\bf] at (6,-1.5) {Ours};
%     \node[font=\bf] at (1.5, -1.5) {4DGS};
%     \node[font=\bf] at (-3,-1.5) {Dyna3D};
%    \end{tikzpicture}
%   \caption{On the Neural 3D Video (N3DV) dataset, our method is qualitatively compared with D3DGS and 4DGS.}
%   \label{fig:fig2-2}
%   \Description{}
% \end{figure*}

\section{Experiments}
\label{sec:experiments}
This section presents both qualitative and quantitative evaluations of dynamic novel view rendering performance using public datasets. We compare the proposed method with state-of-the-art approaches.

\subsection{Implementation Details}
% During the experimental phase, we optimize processing efficiency by downsampling the image resolution by a factor of 2. By using the Adam optimizer with an adaptive learning rate, we are able to accelerate training and improve model performance. In our setup, a single NVIDIA RTX 4090 GPU provides sufficient computational power to handle these downsampled images.
% In our experiment, we leverage COLMAP to accurately reconstruct the geometric structure of a 3D scene from a set of 2D images, including point cloud data and corresponding camera poses. This process provides each model with high-quality initial point cloud data. By employing the Adam optimizer with an adaptive learning rate, we are able to accelerate training and enhance model performance. In our setup, a single NVIDIA RTX 4090 GPU provides sufficient computational power to process these images.
During the experimental phase, we use COLMAP~\cite{schonberger2016structure} to reconstruct the geometric structure of 3D scenes, including point clouds and camera poses, providing each model with high-quality initial point cloud data. By leveraging the Adam optimizer with an adaptive learning rate and a single NVIDIA RTX 4090 GPU, we effectively accelerate training and enhance model performance.
% GPU
% parameters

\subsection{Datasets and Metrics}
\paragraph{Public Datasets.}
% This paper uses two real-world datasets: Neural 3D Video (N3DV)~\cite{an15} and \textcolor{red}{MeetRoom~\cite{an7} }. The N3DV dataset is captured using a 21-camera multi-view system, while the MeetRoom dataset is recorded with a multi-view setup consisting of 13 Azure Kinect cameras. Specifically, the N3DV dataset includes six distinct scenes: \textit{Coffee Martini}, \textit{Cook Spinach}, \textit{Cut Roasted Beef}, \textit{Flame Salmon}, \textit{Flame Steak}, and \textit{Sear Steak}. Each scene contains 300 frames, featuring extended durations and diverse actions, with some scenes incorporating multiple moving objects. 
% %We use the N3DV dataset to evaluate the model's performance in capturing dynamic regions.
% \textcolor{red}{In the collection of the MeetRoom dataset benchmark, the system synchronizes all camera shutters based on external devices. Compared to other commonly used datasets, such as LLFF~\cite{mildenhall2019local} and N3DV~\cite{an15}, the MeetRoom dataset has more sparsely aligned views}.
This study utilizes two real-world datasets: Neural 3D Video (N3DV)~\cite{an15} and the Technicolor Light Field Dataset~\cite{sabater2017dataset}. The N3DV dataset is captured using a multi-view system consisting of 21 cameras, while the Technicolor dataset records video sequences using a $4\times 4$ array of 16 cameras. These cameras are precisely synchronized in time and can capture high-resolution images with a spatial resolution of up to $2048 \times 1088$ pixels.  
Specifically, we select four different scenes from the N3DV dataset: \textit{Cook Spinach, Cut Roasted Beef, Flame Steak, and Sear Steak}. Each scene consists of 300 frames, featuring extended durations and diverse motions, with some scenes containing multiple moving objects.  
For the Technicolor Light Field Dataset, we choose four distinct scenes: \textit{Birthday, Painter, Train, and Fatma}. These scenes not only enhance the dataset's diversity but also provide a comprehensive testing environment for model evaluation.
% \paragraph{\textcolor{red}{Baselines.}}
% For view synthesis across multiple datasets, we conducted comparisons on the N3DV dataset with algorithms such as Mix Voxels\cite{an6}, K-Planes\cite{an19}, HexPlane\cite{an4}, HyperReel\cite{}, NeRFPlayer\cite{an20}, StreamRF\cite{an7}, SWinGS\cite{a10}, D3DGS\cite{a3}, and 4DGS\cite{a2}. On the MeetRoom dataset, we compared our method with STG\cite{a12}, D3DGS\cite{a3}, and 4DGS\cite{a2}. The comparisons were made across multiple metrics.

\begin{table*}[]
    \centering  
    \caption{\textbf{Comparison of methods in novel view rendering based on the N3DV dataset.} Best results are highlighted in \textbf{bold}.}  
    \label{tab:N3DV}  
    \resizebox{\linewidth}{!}{
    \begin{tabular}{lcccccccccccc} \hline  
        \multirow{2}{*}{\centering Method} & \multicolumn{3}{c}{Cook Spinach} & \multicolumn{3}{c}{Sear Steak} & \multicolumn{3}{c}{Flame Steak} & \multicolumn{3}{c}{Cut Roast Beef} \\ 
        \cmidrule(lr){2-4} \cmidrule(lr){5-7}  \cmidrule(lr){8-10} \cmidrule(lr){11-13} 
         & {PSNR$\uparrow$} & {SSIM$\uparrow$} & {LPIPS$\downarrow$} & {PSNR$\uparrow$} & {SSIM$\uparrow$} & {LPIPS$\downarrow$} & {PSNR$\uparrow$} & {SSIM$\uparrow$} & {LPIPS$\downarrow$} & {PSNR$\uparrow$} & {SSIM$\uparrow$} & {LPIPS$\downarrow$}  \\   \midrule 
        MixVoxels~\cite{an16}  & 31.39  & 0.931  & 0.113  & 30.85  & 0.940  & 0.103  & 30.15  & 0.938  & 0.108  & 31.38  & 0.928  & 0.111 \\ 
        K-Planes~\cite{an19} & 31.23  & 0.926  & 0.114  & 30.28  & 0.937  & 0.104  & 31.49  & 0.940  & 0.102 & 31.87  & 0.928  & 0.114 \\ 
        HexPlane~\cite{an4} & 31.05  & 0.928  & 0.114  & 30.00  & 0.939  & 0.105  & 30.42  & 0.939  & 0.104 & 30.83  & 0.927  & 0.115 \\ 
        HyperReel~\cite{an18} & 31.77  & 0.932  & 0.090  & 31.88  & 0.942  & 0.080  & 31.48  & 0.939  & 0.083  & 32.25  & 0.936  & 0.086 \\ 
        NeRFPlayer~\cite{an20} & 30.58  & 0.929  & 0.113  & 29.13  & 0.908  & 0.138  & 31.93  & 0.950  & 0.088  & 29.35  & 0.908  & 0.144 \\ 
        StreamRF~\cite{an7} & 30.89  & 0.914  & 0.162  & 31.60  & 0.925  & 0.147  & 31.37  & 0.923  & 0.152 & 30.75  & 0.917  & 0.154 \\ 
        SWinGS~\cite{a10} & 31.96  & 0.946  & 0.094  & 32.21  & 0.950  & 0.092  & 32.18  & 0.953  & 0.087 & 31.84  & 0.945  & 0.099 \\ 
        D3DGS~\cite{a3} & 20.53  & 0.881  & 0.153  & 25.02  & 0.944  & 0.072  & 23.02  & 0.919  & 0.113 & 22.35  & 0.907  & 0.125 \\ 
        4DGS~\cite{a2} & 28.12  & 0.940  & \textbf{0.038}  & 29.07  & 0.957  & \textbf{0.028}  & 25.04  & 0.918  & 0.079 & 29.71  & 0.944  & \textbf{0.033} \\
        % FSGS~\cite{zhu2025fsgs} & 29.60 & 0.919 & 0.115 & 29.87 &0.945 & 0.116 & 29.42 & 0.943 & 0.113 &28.46 &0.913 &	0.122 \\
        SCGS~\cite{huang2024sc} & 17.20 & 0.734 & 0.232&  28.77 & 0.951 & 0.056&  23.49 &0.902 & 0.104& 6.29 & 0.007& 0.683 \\
        Ours & \textbf{32.59}  & \textbf{0.966}  & 0.054   & \textbf{33.12}  & \textbf{0.973}  & 0.049  & \textbf{33.34}  & \textbf{0.971}  & \textbf{0.052} &\textbf{33.06}  & \textbf{0.969}  & 0.055 \\ 
        \midrule

     \end{tabular} }
\end{table*}

\begin{table*}%[h]
    \centering  
    \caption{\textbf{Methods comparison on the Technicolor dataset.} Best results are highlighted in \textbf{bold}. }  
    \label{tab:Technicolor}  
    \resizebox{\linewidth}{!}{
    \begin{tabular}{lcccccccccccc} \hline  
        \multirow{2}{*}{\centering Method} & \multicolumn{3}{c}{Birthday} & \multicolumn{3}{c}{Painter} & \multicolumn{3}{c}{Train} & \multicolumn{3}{c}{Fatma} \\ 
        \cmidrule(lr){2-4} \cmidrule(lr){5-7} \cmidrule(lr){8-10}  \cmidrule(lr){11-13} 
         & {PSNR$\uparrow$} & {SSIM$\uparrow$} & {LPIPS$\downarrow$} & {PSNR$\uparrow$} & {SSIM$\uparrow$} & {LPIPS$\downarrow$} & {PSNR$\uparrow$} & {SSIM$\uparrow$} & {LPIPS$\downarrow$} & {PSNR$\uparrow$} & {SSIM$\uparrow$} & {LPIPS$\downarrow$}  \\   \midrule  

        D3DGS\cite{a3} & 33.81  & 0.965  & 0.014  & 37.38  & 0.957  & 0.036 & - &- & -& 38.40 &	0.911 &	0.093  \\ 
        STG\cite{a12} & 33.87  & 0.951  & 0.038  & 37.30  & 0.928  & 0.095  & 33.36 &	0.948 &	0.036 & 37.28 	& 0.906 &	0.155 \\
        FSGS\cite{zhu2025fsgs} & 26.26  & 0.920  & 0.068  & 34.36  & 0.958  & 0.063  & 30.39 &	0.965 &	0.032 & 27.62 	& 0.825 &	0.276 \\
        4DGS\cite{a2} & 21.94  & 0.902  & 0.071  & 28.61  & 0.940  & 0.058  & 22.36 &	0.878 &	0.124 & 23.42 &0.763 &	0.236 \\
        Ours & \textbf{34.72}  & \textbf{0.988}  & \textbf{0.013}  & \textbf{38.37}  & \textbf{0.985}  & \textbf{0.022}& \textbf{35.30} &	\textbf{0.990} &	\textbf{0.008} & \textbf{38.91} 	&\textbf{0.945} 	&\textbf{0.071} \\
        \midrule
     \end{tabular} }
\end{table*}

\paragraph{Metrics.}
To evaluate the novel view rendering performance of our models, we use the following three metrics in the experimental section: Peak Signal-to-Noise Ratio (PSNR), Structural Similarity Index (SSIM), and Learned Perceptual Image Patch Similarity (LPIPS)~\cite{zhang2018unreasonable}. These three metrics offer different perspectives for assessing the quality of generated images. Specifically, PSNR and SSIM evaluate image quality on a pixel-wise and structural basis, respectively, while LPIPS compares deep features extracted by AlexNet~\cite{krizhevsky2012imagenet} to assess perceptual similarity between two images.

% The results are shown in Tables 1-3. 
% We also performed ablation experiments on various aspects of the method. 

\subsection{Novel View Rendering on N3DV}
% As shown in Table~\ref{tab:N3DV}, our method outperforms other approaches and achieve the best performance, especially in the LPIPS metric. 
As shown in Table~\ref{tab:N3DV}, our method outperforms others overall, achieving relatively strong performance, with the best PSNR and SSIM scores on each sub-dataset.
% Additionally, we calculated the average PSNR and SSIM scores across all scenes and presented them in a table, as shown in Figure \ref{fig:method_comparison}. Our proposed framework demonstrates impressive performance on average across all scenes.
The Mixvoxel algorithm~\cite{an16} converts point cloud data into a regular voxel grid, simplifying the data processing pipeline. This voxel representation allows for efficient feature extraction. However, during the voxelization process, some fine-grained details may be lost due to the conversion of point cloud data into a regular grid. As a result, the algorithm's performance may be affected in complex scenarios. Moreover, the performance of the Mixvoxel algorithm is sensitive to voxel parameters, such as voxel size and resolution. Different parameter settings can lead to significant fluctuations in performance, requiring careful adjustment and optimization. This adds complexity and challenges in applying the algorithm to practical tasks.

K-Planes~\cite{an19} and HexPlane~\cite{an4} achieve accelerated rendering by storing information in feature grids. While this approach significantly improves rendering speed, the grid-based representation fails to adequately adapt to dynamic scene changes, particularly in the case of fast-moving objects. In contrast, NeRFPlayer~\cite{an20} introduces innovative methods for handling dynamic scenes, offering enhanced adaptability to scene variations. However, these approaches still face challenges when dealing with large viewpoint changes or highly dynamic scenes, leading to a degradation in rendering quality or a reduction in computational efficiency.

Streaming Radiance Fields (StreamRF)~\cite{an7} employs an explicit grid representation; however, its reliance on online training for dynamic scenes renders it inadequate for accommodating substantial viewpoint shifts and complex dynamic scenarios. Conversely, Sliding Windows for Dynamic 3DGS (SWinGS)~\cite{a10} leverages multi-resolution hash encoding, yet it falls short in capturing high-frequency scene details in multi-view and complex dynamic tasks. Furthermore, its insufficient utilization of depth information results in inconsistent visual outputs.

The Few-shot View Synthesis using Gaussian Splatting (FSGS)~\cite{zhu2025fsgs} method enables real-time, photo-realistic view synthesis with as few as three training views. However, it encounters challenges in capturing fine texture details.
The 4DGS method~\cite{a2} incorporates spatiotemporal properties using the six-plane technique but struggles with processing multiple perspectives effectively. Although D3DGS~\cite{a3} employs deformation fields to account for dynamic changes, it treats frames as discrete samples, adapting to time-dependent trajectories or deformations while neglecting the rich motion cues available from continuous two-dimensional observations.

%The 4DGS method considers the spatiotemporal properties through the six-plane method, but it is less effective in processing multiple perspectives. Although D3DS uses deformation fields to consider dynamic changes, they treat frames as discrete samples to adapt to time-related trajectories or deformations, while ignoring the rich motion clues under continuous two-dimensional observations.

In contrast, our proposed method not only accounts for the discrete nature of motion frames but also captures the continuity between frames through learnable infinite Taylor series. Additionally, we establish a rigid connection between adjacent Gaussian points, ensuring spatiotemporal consistency across points.

As shown in Figure \ref{fig:fig2-1}, the comparison of reconstruction effects demonstrates that our method produces clearer and more faithful images than the other models. 
% Although Mixvoxel~\cite{an16} delivers good reconstruction results on the N3DV dataset, its performance is poor on MeetRoom, indicating that it is sensitive to parameters and unstable across different datasets. This suggests that Mixvoxel~\cite{an16} requires re-optimization for each specific dataset. 
The 4DGS~\cite{a2} method, while effective, may demand more computational resources and time when processing large-scale dynamic scenes or performing high-resolution rendering. Its performance is highly dependent on the accuracy and completeness of the input data; noisy or incomplete input can negatively impact both the modeling and rendering quality. Finally, D3DGS~\cite{a3} shows blurred motion areas when rendering dynamic scenes, indicating that there is considerable room for improvement in its ability to capture and render dynamic motion accurately. Further details of the experimental comparison and analysis are provided in the supplementary material.

%The Comparison of reconstruction effects, As shown in Figures 1 to 3, can be seen that our method reconstructs clearer and more fidelity images than other models. Although Mixvoxel has a good reconstruction effect on N3DV, it has a poor reconstruction effect on meetroom. It can be seen that its model performance is unstable and sensitive to parameters, and needs to be re-optimized for different data sets. 4DGS may require more computing resources and time when processing large-scale dynamic scenes or requiring high-resolution rendering. The performance of 4DGS depends largely on the accuracy and completeness of the input data. If the input data is noisy or missing, it may affect its modeling effect and rendering quality.D3Dgs has blurred motion areas when rendering dynamic scenes, which shows that there is still a lot of room for improvement in its ability to capture dynamic scenes.

\subsection{Novel View Rendering on Technicolor Dataset}
Here, we conduct a detailed comparison against state-of-the-art approaches to further validate the superiority of our method. By analyzing multiple sequences within the dataset, we highlight the consistency and robustness of our approach in handling diverse and complex scenes. The following quantitative results illustrate the substantial improvements achieved by our method over existing techniques.

As shown in Table~\ref{tab:Technicolor}, the proposed method yields significantly more accurate and robust results. 
% Specifically, our model excels in PSNR, SSIM, and LPIPS, demonstrating its effectiveness in accurately evaluating performance across various scenarios. 
For example, in the \textit{Birthday} sequence, the PSNR of the proposed method is 34.72, whereas the state-of-the-art methods 4DGS~\cite{a2}, FSGS~\cite{zhu2025fsgs}, D3DGS~\cite{a3}, and STG~\cite{a12} achieve PSNR values of 21.94, 26.26, 33.81, and 33.87, respectively. This results in improvements of approximately $\mathbf{58.25\%}$, $\mathbf{32.22\%}$, $\mathbf{2.69\%}$, and $\mathbf{2.51\%}$ over these methods. Similar trends are observed in other sequences, including \textit{Painter}, \textit{Train}, and \textit{Fatma}. Compared to STG~\cite{a12} and 4DGS~\cite{a2}, the D3DGS~\cite{a3} method demonstrates more robust performance across sequences, particularly in the \textit{Painter}, \textit{Fatma} sequence, where it achieves the best PSNR score and also delivers competitive LPIPS and SSIM results.

% As listed in Table~\ref{tab:meetroom}, the proposed method has achieved much more accurate and robust results. For example, in the Sequence of \textit{Discussion}, the PSNR of the proposed method is $36.77$, while the corresponding results of state-of the-art methods, STG~\cite{}, D3DGS~\cite{}, and 4DGS~\cite{} are $25.53$, $32.52$, and $30.61$, respectively. Therefore, the improvement of the proposed method in this seuqence compared to these three methods are \textcolor{red}{$30\%$,  $30\%$, and $30\%$}, respectively. Furthermore, the similar phonomen can be witnessed in other sequences, including in the \textit{Vrheadset}, \textit{Trimming} and \textit{Stepin}. Compared to STG~\cite{} and D3DGS~\cite{}, the 4DGS method shows more robust performance in sequences, espcially in the \textit{Stepin} sequence, its LPIPS result is the best value than other methods and its PSNR and SSIM performance also achieved acceptable performance.     

\subsection{\textcolor{black}{Ablation Study}}

 We conducts ablation experiments on several proposed parts, as shown in the Table \ref{tab:ablation_study_N3DV}. In our ablation experiments, we configured the settings as follows: \textit{w/o Time-opacity}, \textit{w/o Time-motion}, \textit{w/o Time-rotation}, and so on. Results indicate a significant drop in model performance when a time-varying mathematical model is not constructed, highlighting the importance of \textit{Time-opacity}, \textit{Time-motion}, \textit{Time-rotation}, and \textit{Time-scale} in our framework. 
 Furthermore, we conducted an in-depth analysis of the Peano remainder and observed that modeling the learnable infinite Taylor series for all Gaussian points without accounting for higher-order terms using the Peano remainder results in performance degradation. This finding underscores the importance of the Peano remainder in constructing the infinite Taylor series of Gaussian points. By leveraging the Peano remainder, we effectively control the model's approximation error, thereby achieving improved accuracy and stability in the 3D reconstruction of dynamic scenes.
% Furthermore, we conducted an in-depth analysis of the Peano remainder and found that directly modeling the motion equations for all Gaussian points without using the Peano remainder to account for the influence of higher-order terms leads to a degradation in model performance. This finding further confirms the crucial role of the Peano remainder in constructing the motion equations of Gaussian points, as it is essential for enhancing the overall model performance. By effectively utilizing the Peano remainder, we are able to control the approximation error of the model, achieving higher accuracy and stability in the 3D reconstruction of dynamic scenes.

% \textcolor{red}{Additionally, we conducted an in-depth analysis of sampling points. We observed that if the motion field training is applied directly to all Gaussian points without initial pre-training on a subset of sampled points, the model’s performance also declines. This finding further confirms the critical role of pre-training the motion field on a subset of points to enhance overall model performance.}

\begin{table}[h]
    \centering  
    \caption{\textbf{Ablation study on the N3DV dataset.} best results are highlighted in \textbf{bold}.}  \small
    \resizebox{\linewidth}{!}{
    \begin{tabular}{lccccccccc} \hline  
        Method & {PSNR$\uparrow$} & {SSIM$\uparrow$} & {LPIPS$\downarrow$} \\
            \midrule  
            
        w/o Time-opacity  & 31.17  & 0.952  & 0.096  \\ 
        w/o Time-motion  & 29.24  & 0.920  & 0.154  \\
        w/o Time-rotation  & 31.21  & 0.953  & 0.103  \\
        w/o Time-scale & 31.40  & 0.953  & 0.097   \\ 
        w/o Peano remainder & 31.51  & 0.935  & 0.103  \\ 
        Ours Full  & \textbf{33.03} & \textbf{0.970} &	\textbf{0.052}  \\ \hline
     \end{tabular} }
      \label{tab:ablation_study_N3DV}
\end{table}

\section{Discussion and Conclusion}
In this paper, we address the challenge of capturing the time-dependent properties (position, rotation, and scale) of Gaussians in dynamic scenes. The vast number of time-varying Gaussian parameters, coupled with the constraints imposed by limited photometric data, complicates the task of finding an optimal solution. While end-to-end neural networks offer a promising approach by learning complex temporal dynamics directly from data, they suffer from the lack of explicit supervision and often fail to produce high-quality transformation fields. On the other hand, time-conditioned polynomial functions provide a more explicit solution for modeling Gaussian trajectories and orientations, but their effectiveness is limited by the need for extensive handcrafted design and the difficulty of developing a generalizable model across diverse scenes.

To overcome these limitations, we propose a novel method based on a learnable infinite Taylor series. This approach combines the strengths of both implicit neural representations and explicit polynomial approximations, enabling accurate modeling of the dynamic behavior of Gaussians over time. Our method is shown to outperform existing approaches in both qualitative and quantitative multi-view evaluations, offering a more robust and flexible solution for dynamic scene reconstruction. There are several directions for future research. One potential avenue is the extension of the approach to handle more complex and highly dynamic scenes, where the current model may need further refinement to maintain accuracy and robustness.

\clearpage

\section{Acknowledgments}
The research/project is supported by the National Research Foundation (NRF)
Singapore, under its NRF-Investigatorship Programme (Award ID. NRF-NRFI09-0008), the National Natural Science Foundation of China (No.62072388), Fujian Provincial Science and Technology Major Project (No.2024HZ022003), Jiangxi Provincial Natural Science Foundation Key Project (No.20244BAB28039), Xiamen Public Technology Service Platform (No.3502Z20231043), and Fujian Sunshine Charity Public Welfare Foundation.
%s\clearpage
%\setcounter{page}{1}
%\maketitlesupplementary
\appendix
% \setcounter{page}{1}
% \setcounter{table}{3}
% \setcounter{figure}{2}

% \section{Rationale}
% \label{sec:rationale}
% % 
% Having the supplementary compiled together with the main paper means that:
% % 
% \begin{itemize}
% \item The supplementary can back-reference sections of the main paper. For example, we can refer to \cref{sec:intro};
% \item The main paper can forward reference sub-sections within the supplementary explicitly (e.g., referring to a particular experiment); 
% \item When submitted to arXiv, the supplementary will already be included at the end of the paper.
% \end{itemize}
% % 
% To split the supplementary pages from the main paper, you can use \href{https://support.apple.com/en-ca/guide/preview/prvw11793/mac#:~:text=Delete%20a%20page%20from%20a,or%20choose%20Edit%20%3E%20Delete).}{Preview (on macOS)}, \href{https://www.adobe.com/acrobat/how-to/delete-pages-from-pdf.html#:~:text=Choose%20%E2%80%9CTools%E2%80%9D%20%3E%20%E2%80%9COrganize,or%20pages%20from%20the%20file.}{Adobe Acrobat} (on all OSs), as well as \href{https://superuser.com/questions/517986/is-it-possible-to-delete-some-pages-of-a-pdf-document}{command line tools}.
% \section{}

\definecolor{lightred}{rgb}{1, 0.4, 0.4}

\section{Novel View Rendering}
% In the evaluation section, we have designed a series of comprehensive experiments to assess the performance of our method. Here, we further provide a set of \textbf{quantitative and qualitative} results to more thoroughly validate the effectiveness of our approach.
In the evaluation section, we have designed a series of comprehensive experiments to assess the performance of our method. Here, we present a set of \textbf{visual results} to further validate the effectiveness of our approach more thoroughly.

\subsection{Qualitative Analysis of Details}
In the quantitative analysis of novel view rendering algorithms, we focused on several key evaluation metrics, including Peak Signal-to-Noise Ratio (PSNR), Structural Similarity Index (SSIM) and Perceptual Image Patch Similarity (LPIPS)~\cite{zhang2018unreasonable}. These metrics help us quantify the details and overall quality of the reconstructed images. By comparing these metrics, 
% as shown in Table~\ref{tab:N3DV} and Table~\ref{tab:meetroom}, 
we can more accurately assess the performance differences between different algorithms. To gain a more comprehensive understanding, we also incorporated qualitative analysis, examining the detailed performance of various algorithms in the reconstructed images, leading to a deeper evaluation. Through visual presentation, we can further assess the strengths and weaknesses of the algorithms, ensuring a multidimensional and comprehensive understanding of their performance.

\begin{figure*}[h]
  \centering
  \includegraphics[width=\textwidth]{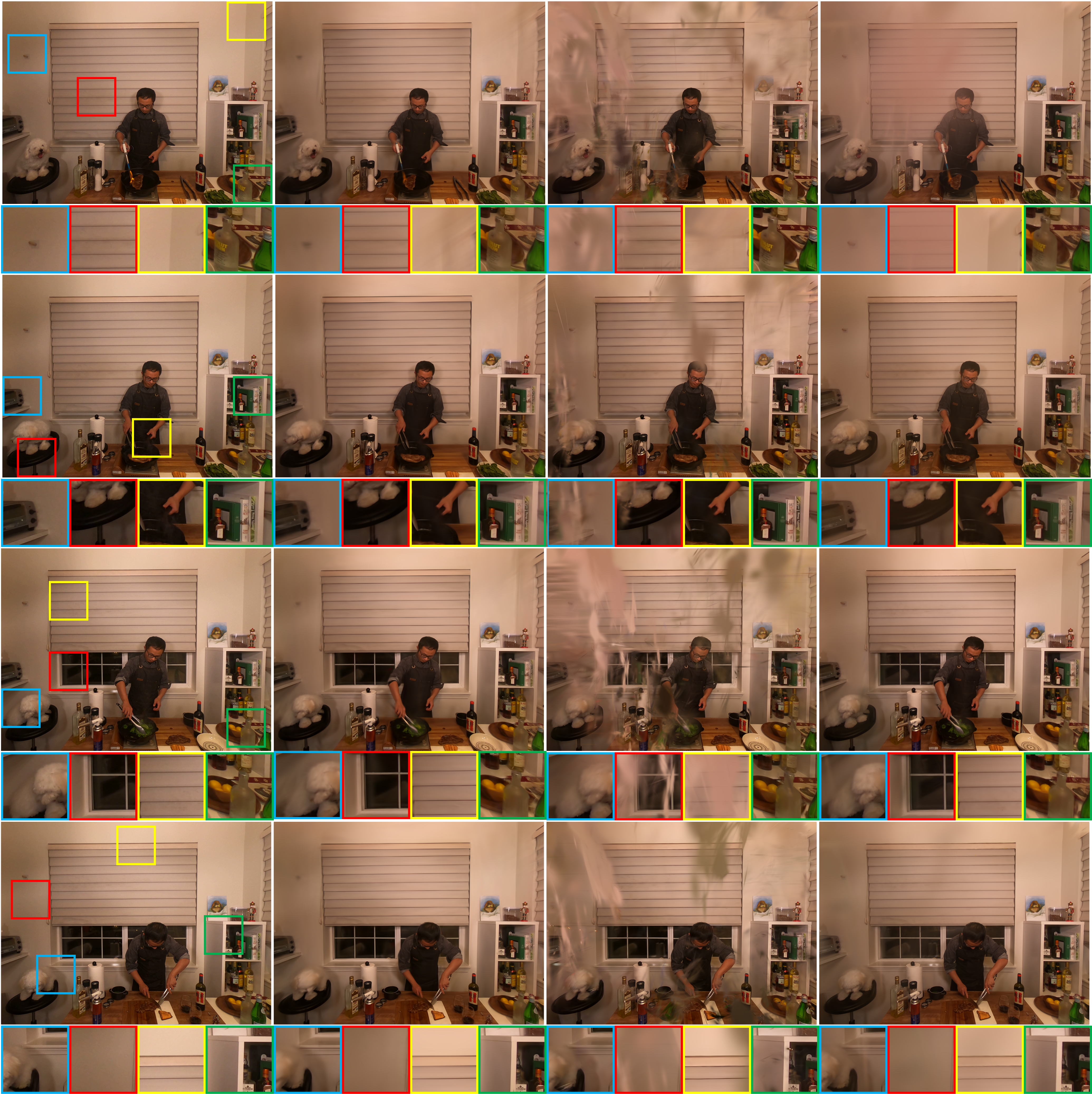} % 替换为你的图像文件名
    \begin{tikzpicture}
    \node[font=\bf] at (11,-1.5) {};
    \node[font=\bf] at (10,-1.5) {4DGS~\cite{a2}};
    \node[font=\bf] at (6,-1.5) {D3DGS~\cite{a3}};
    \node[font=\bf] at (1.5, -1.5) {Ours};
    \node[font=\bf] at (-3,-1.5) {GT};
  \end{tikzpicture}
  \caption{Qualitative analysis of novel view rendering on the N3DV dataset, comparing the detail information of reconstructed images from different algorithms. }
  \label{fig:figA-1}
\end{figure*}

% \begin{figure*}[h]
%   \centering
%   \includegraphics[width=\textwidth]{construc_detail3_2.png} % 替换为你的图像文件名
%   \includegraphics[width=\textwidth]{construc_detail3_4.png} % 替换为你的图像文
%   \includegraphics[width=\textwidth]{construc_detail3_1.png} % 替换为你的图像文
%   \includegraphics[width=\textwidth]{construc_detail3_3.png} % 替换为你的图像文
%   \begin{tikzpicture}
%     \node[font=\bf] at (11,-1.5) {};
%     \node[font=\bf] at (10,-1.5) {GT};
%     \node[font=\bf] at (6,-1.5) {Ours};
%     \node[font=\bf] at (1.5, -1.5) {4DGS~\cite{a2}};
%     \node[font=\bf] at (-3,-1.5) {D3DGS~\cite{a3}};
%   \end{tikzpicture}
%   \caption{Qualitative analysis of novel view rendering on the N3DV dataset, comparing the detail information of reconstructed images from different algorithms. }
%   \label{fig:figA-1}
% \end{figure*}

As shown in Figure~\ref{fig:figA-1}, we can see that our algorithm demonstrates superior performance in detail reconstruction compared to others. However, D3DGS and 4DGS face challenges such as artifacts and distortions during the reconstruction process. We provide a detailed explanation of each row in Figure~\ref{fig:figA-1}:

% \textit{First Row}: D3DGS and 4DGS struggle with blurry reconstructions in areas with fine details, accompanied by significant noise. In scenes with glass occlusion, the recovery of intricate details is insufficient, further diminishing reconstruction quality.  

% \textit{Second Row}: Texture reconstruction by D3DGS and 4DGS is subpar, evident in blurry black dots and the incomplete restoration of the green bottle. In complex scenarios with multiple objects, such as the region highlighted in the green box, the reconstruction between bottles reveals significant flaws.  

% \textit{Third Row}: Both methods perform poorly in occluded scenes. For instance, glass occlusion hampers the accurate reconstruction of scenes behind the glass. Similarly, bottle occlusion and glass refraction cause deformation and blurriness in the reconstruction of the orange object.  

% \textit{Fourth Row}: D3DGS and 4DGS exhibit color distortion, particularly under indoor lighting conditions. Additionally, certain areas, such as the edges of the pan and text, appear blurry. The images lack contrast; for example, in the region marked by the red box, reflective textures and hair details are inaccurately reconstructed.  

% **Flame Steak**  
\textit{First Row}: Overall, the texture reconstruction quality of D3DGS and 4DGS is below the standard. Additionally, as seen in the red box (curtain reconstruction) and the blue box (wall reconstruction), the detail reconstruction performance of D3DGS and 4DGS is also poor.  

% **Sear Steak**  
\textit{Second Row}: The overall reconstruction performance of D3DGS is poor. Both 4DGS and D3DGS exhibit issues in detail reconstruction, such as blurred shadows in the yellow box, significant reflections and artifacts on the leather stool in the red box, and additional artifacts appearing in the blue box for D3DGS.  

% **Cook Spinach**  
\textit{Third Row}: D3DGS performs poorly in both overall reconstruction quality and detail representation. 4DGS also has some issues in detail reconstruction, such as unexplained black spots above the white bottle in the green box and unexplained light appearing on the left side of the blue box.  

% **Cut Roasted Beef**  
\textit{Fourth Row}: Both D3DGS and 4DGS exhibit color distortions. Additionally, shadows appear in certain areas (e.g., red, yellow, and blue boxes), and the image in the blue box lacks contrast. There are also extraneous elements in the green box of D3DGS.

% \begin{figure*}[htbp]
%   \centering
%   \includegraphics[width=\textwidth]{construc_detail2_2.png} % 替换为你的图像文件名
%   \includegraphics[width=\textwidth]{construc_detail2_4.png} % 替换为你的图像文
%   \includegraphics[width=\textwidth]{construc_detail2_1.png} % 替换为你的图像文
%   \includegraphics[width=\textwidth]{construc_detail2_3.png} % 替换为你的图像文
%   \begin{tikzpicture}
%     \node[font=\bf] at (11,-1.5) {};
%     \node[font=\bf] at (10,-1.5) {GT};
%     \node[font=\bf] at (6,-1.5) {Ours};
%     \node[font=\bf] at (1.5, -1.5) {4DGS~\cite{a2}};
%     \node[font=\bf] at (-3,-1.5) {D3DGS~\cite{a3}};
%   \end{tikzpicture}
%   \caption{Qualitative analysis of novel view rendering on the N3DV dataset, comparing the detail information of reconstructed images from different algorithms.}
%   \label{fig:figA-2}
% \end{figure*}

\subsection{Qualitative Analysis of Ablation Experiments}

% **A Qualitative Analysis of Ablation Experiments on the Cut Roasted Beef Class in the N3DV Dataset for Novel View Rendering**

In our ablation experiments on the \textit{Sear Steak} class in the N3DV dataset, we conducted an in-depth qualitative analysis to evaluate the impact of ablating different modules on the performance of reconstructed images and the representation of fine details. By systematically comparing images reconstructed after the ablation of various modules, we were able to uncover their respective strengths and limitations in handling complex scenes.

First, significant differences were observed in rendering quality across the ablations of different modules. Ablating specific modules reduced the ability to capture geometric details of objects, as shown in Figure~\ref{fig:figA-2}, such as surface textures and edge contours. 
For example, \textcolor{blue}{as highlighted by the blue bounding box}, both \textit{w/o Peano remainder} and \textit{w/o Time-opacity} failed to accurately capture geometric details, leading to missing geometric information. Similarly, \textcolor{green}{as shown in the green bounding box}, \textit{w/o Peano remainder}, \textit{w/o Time-opacity}, and \textit{w/o Time-scale} exhibited poor performance in reconstructing surface textures, producing artifacts such as shadowing and linear streaks. Additionally, \textit{w/o Peano remainder} and \textit{w/o Time-opacity} demonstrated a weaker capability in capturing edge contours, resulting in blurred or muddled details during reconstruction.  

In other cases, module ablations introduced noticeable noise or over-smoothing in specific details.  
For instance, \textcolor{red}{as illustrated in the red bounding box}, \textit{w/o Time-motion} and \textit{w/o Time-rotation} introduced significant noise when reconstructing fine details compared to the original images. These differences were particularly pronounced when processing \textit{Sear Steak} samples with rich geometric features, highlighting the critical role of these modules in maintaining reconstruction fidelity.  
Furthermore, we evaluated the impact of different module ablations on handling complex scenes. \textcolor{yellow}{As shown in the yellow bounding box}, the reconstruction quality of \textit{w/o Time-motion}, \textit{w/o Time-rotation}, and \textit{w/o Time-scale} was relatively blurry, with increased noise and excessive smoothing, ultimately degrading the overall visual quality. This underscores the importance of these modules in accurately capturing fine details in complex scenes.
% # Albation figure
\begin{figure*}[h]
  \centering
  \includegraphics[width=\textwidth]{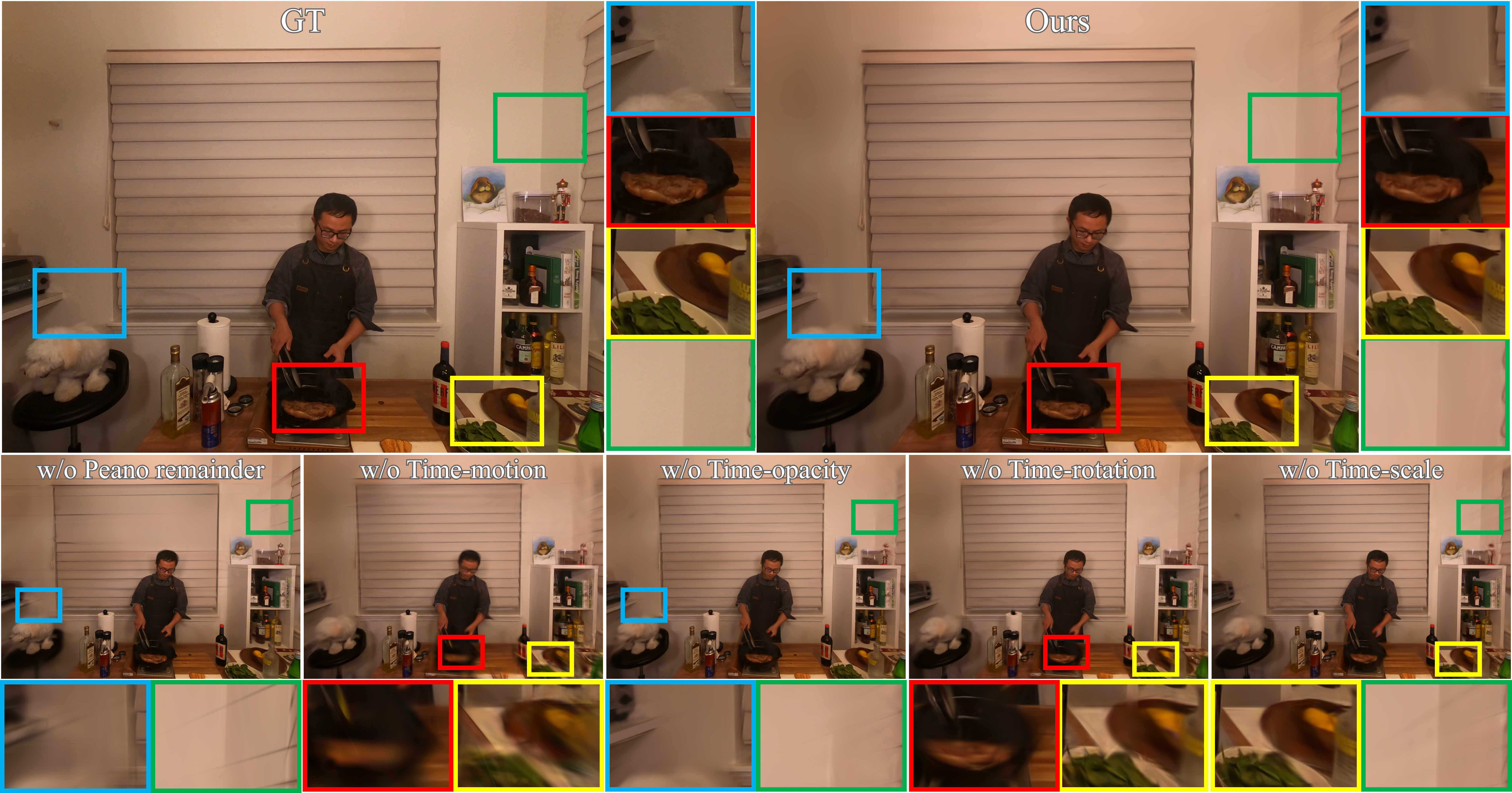} % 替换为你的图像文件名

  \caption{\textit{Sear Steak} Novel View Rendering on the N3DV Dataset: Qualitative Analysis of Ablation Experiments - Comparison of Reconstruction Quality and Detail Representation with Module Ablations. }
  \label{fig:figA-2}
\end{figure*}

% \begin{table}%[h]
%     \centering  
%     \caption{\textbf{Methods comparison in Technicolor dataset.}}  
%     \label{tab:Technicolor}  
%     \resizebox{\linewidth}{!}{
%     \begin{tabular}{lcccccc} \hline  
%         \multirow{2}{*}{\centering Method} & \multicolumn{3}{c}{Birthday} & \multicolumn{3}{c}{Painter} \\ 
%         \cmidrule(lr){2-4} \cmidrule(lr){5-7} 
%          & {PSNR$\uparrow$} & {SSIM$\uparrow$} & {LPIPS$\downarrow$} & {PSNR$\uparrow$} & {SSIM$\uparrow$} & {LPIPS$\downarrow$}   \\   \midrule  
%         D3DGS\cite{a3} & 33.81  & 0.965  & 0.014  & 37.38  & 0.957  & 0.036  \\ 
%         STG\cite{a12} & 33.87  & 0.951  & 0.038  & 37.30  & 0.928  & 0.095  \\
%         FSGS\cite{zhu2025fsgs} & 26.26  & 0.920  & 0.068  & 34.36  & 0.958  & 0.063  \\
%         4DGS\cite{a2} & 21.94  & 0.902  & 0.071  & 28.61  & 0.940  & 0.058  \\
%         Ours & \textbf{34.72}  & \textbf{0.988}  & \textbf{0.013}  & \textbf{38.37}  & \textbf{0.985}  & \textbf{0.022} \\
%         \midrule
%      \end{tabular} }
% \end{table} 

In summary, this qualitative analysis not only revealed the impact of ablating specific modules on reconstruction quality but also provided deeper insights into their effectiveness in capturing fine details. These findings hold significant implications for optimizing novel view rendering algorithms and improving image quality. By identifying the strengths and limitations of each module, we can better target algorithmic improvements to achieve more accurate and high-quality novel view rendering outcomes. 
\begin{figure*}[h]
  \centering
      \includegraphics[width=\textwidth]{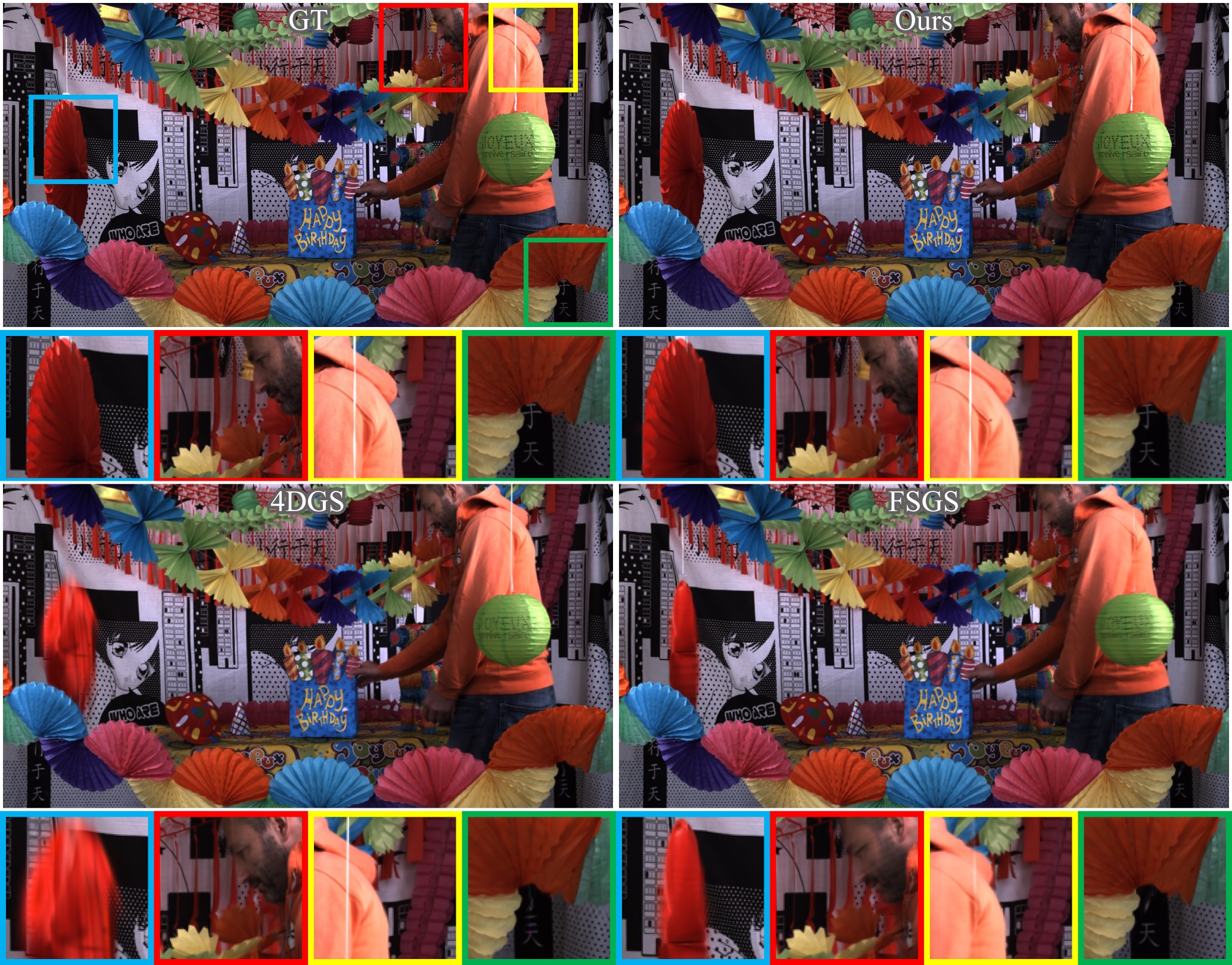} % 替换为你的图像文件名
  \caption{Qualitative analysis of novel view rendering on the Birthday dataset from the Technicolor, comparing the detailed reconstructions of different algorithms.}
  \label{fig:figA-3}
\end{figure*}
\begin{figure*}[h]
  \centering
      \includegraphics[width=\textwidth]{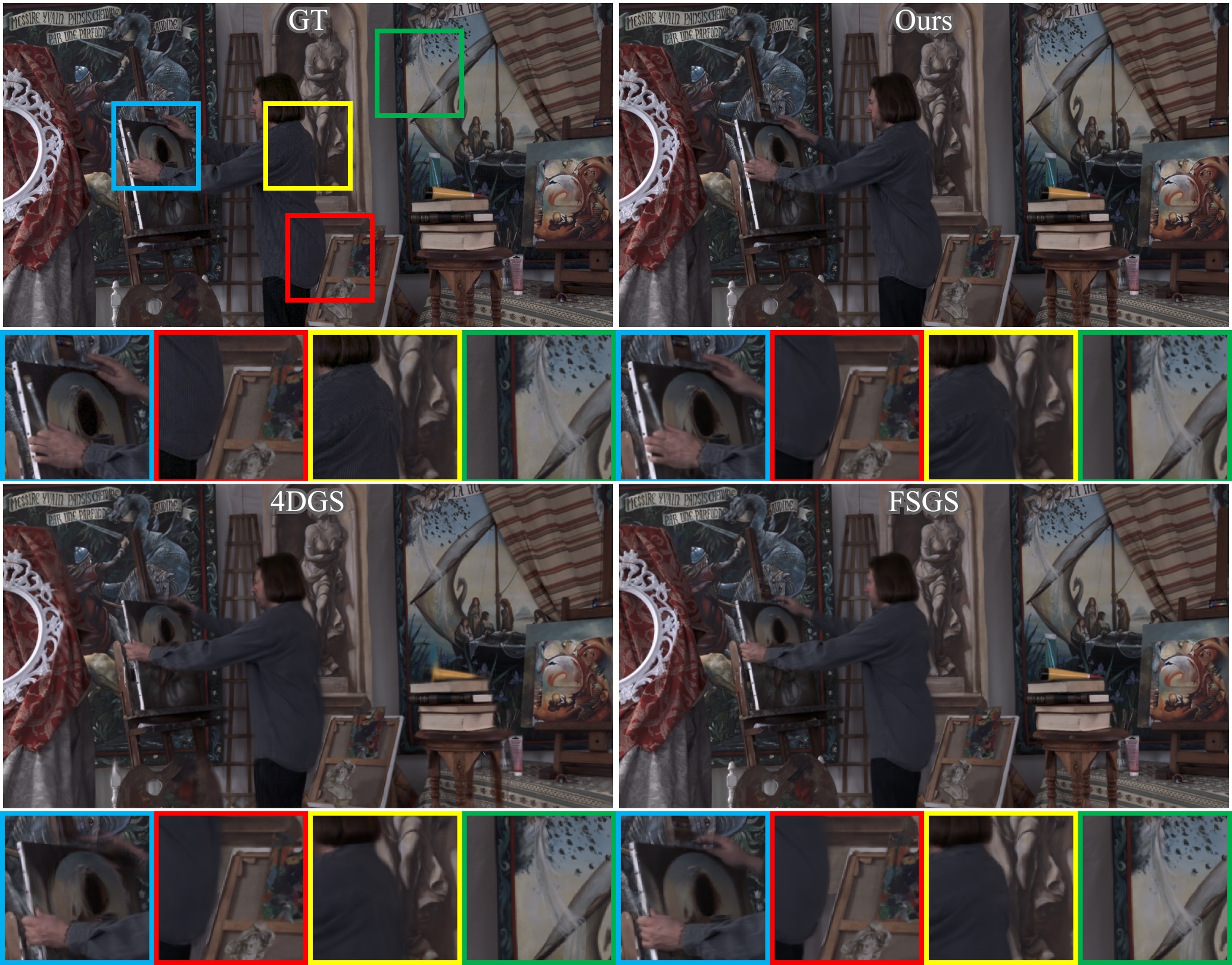} % 替换为你的图像文件名
  \caption{Qualitative analysis of novel view rendering on the Painter dataset from the Technicolor, comparing the detailed reconstructions of different algorithms.}
  \label{fig:figA-4}
\end{figure*}

\section{Performance Analysis of Large-Scale Data Scene Reconstruction}
% \textcolor{red}{To comprehensively evaluate the performance of our model, we conducted tests using the Technicolor Light Field Dataset, a large-scale and high-quality visual resource. This dataset captures video sequences through a 4X4 array of 16 cameras that are precisely synchronized in time, enabling the acquisition of high spatial resolution images with a resolution of up to 2048X1088 pixels. The Technicolor Light Field Dataset encompasses five unique scenes: \textit{Birthday}, \textit{Painter}, \textit{Theater}, \textit{Train}, and \textit{Fabien}. These scenes not only enhance the dataset's diversity but also provide a broad testing environment for model evaluation. We selected two scenes, \textit{Birthday} and \textit{Painter}, for quantitative and qualitative analysis.}
% Our evaluation approach integrates both quantitative and qualitative dimensions. For quantitative analysis, we employed a series of standardized metrics, including PSNR, SSIM, and LPIPS, to objectively measure the model's performance across different scenarios. These metrics offer a clear perspective on the model's capabilities in various testing conditions. For qualitative analysis, we delved deeper into the model's visual performance when handling complex scenes. This included assessing the model's ability to capture details and reconstruct object surface textures. By visually comparing the model's outputs with real-world scenes, we gained a deeper understanding of the model's potential and limitations in practical applications.

To better evaluate the performance of Large-Scale Data Scene Reconstruction, we conducted a qualitative analysis on the Technicolor Light Field Dataset.  
For a more in-depth assessment, we explored the model's visual performance in handling complex scenes, focusing on its ability to capture fine details and reconstruct object surface textures. By visually comparing the model's outputs with real-world scenes, we gained deeper insights into its strengths and limitations in practical applications.

% As shown in Table \ref{tab:Technicolor}, our algorithm outperforms other methods, delivering superior results across multiple standardized metrics. Specifically, our model excels in PSNR, SSIM, and LPIPS, demonstrating its effectiveness in accurately evaluating performance across various scenarios. To vividly illustrate the differences between our method and others, we visualized the qualitative results. 

As highlighted in the boxed regions, it is evident that our method can render higher-quality images. As shown in Figure~\ref{fig:figA-3},~\ref{fig:figA-4}, we can see that in the Birthday scene, our reconstruction captures details better compared to other models. Several issues are observed in the reconstructions of 4DGS and FSGS: \textcolor{blue}{in the area marked by the blue box, both methods exhibit reconstruction blurriness}; \textcolor{red}{in the area marked by the red box, neither 4DGS nor FSGS successfully reconstructs the yellow object near the person's nose bridge, and the images generated by both methods have relatively lower resolution. Additionally, FSGS introduces motion blur artifacts}. \textcolor{yellow}{In the area marked by the yellow box, both methods make errors in reconstruction, mistakenly generating a red object beneath the green leaf}.~\textcolor{green}{ Lastly, in the area marked by the green box, both 4DGS and FSGS fail to accurately reconstruct the text along the edges}.

In the Painter scene, it is evident that our model outperforms other models in reconstruction quality, while both 4DGS and FSGS exhibit the following issues: \textcolor{blue}{in the area marked by the blue box, noticeable hand deformation occurs}; \textcolor{red}{in the area marked by the red box, significant errors are observed in reconstructing the distance between the clothing and surrounding objects}; \textcolor{yellow}{in the area marked by the yellow box, the clothing texture shows clear differences compared to GT}; \textcolor{green}{and in the area marked by the green box, the highlights of the painting are not accurately reconstructed}.

% \section{How Are Local Gaussian Primitives (LPs) Points Estimated Through Global Gaussian Primitives (GPs) Points}
% \section{Theoretical Analysis of Linear Blend Skinning (LBS)}

%  % \input{sec/2_formatting}
%  % \input{sec/3_finalcopy}
% \clearpage

{
    \small
    \bibliographystyle{ieeenat_fullname}
    \bibliography{main}
}

\end{document}